# Object Class Detection and Classification using Multi Scale Gradient and Corner Point based Shape Descriptors

Fernando Basura, Karaoglu Sezer*, Saha Sajib Kumar [†]
Faculty of Science, University Jean Monnet, France
basuraf@gmail.com, sezerkaraolu@yahoo.com*, to_sajib_cse@yahoo.com [†]

*Abstract*—This paper presents a novel multi scale gradient and a corner point based shape descriptors. The novel multi scale gradient based shape descriptor is combined with generic Fourier descriptors to extract contour and region based shape information. Shape information based object class detection and classification technique with a random forest classifier has been optimized. Proposed integrated descriptor in this paper is robust to rotation, scale, translation, affine deformations, noisy contours and noisy shapes. The new corner point based interpolated shape descriptor has been exploited for fast object detection and classification with higher accuracy.

*Key words-gradient based descriptor; corner based descriptor; steerable filter; multi-scale edge response*

## I. INTRODUCTION

According to the review done by Zhang [9], shape descriptors are classified into two classes; contour based and region based. Curvature scale space methods (CSS) [14], [17], [18], global shape descriptors, geometric invariants and spectral descriptors are based on contour information of the shape while Geometric moments, Legendre moments, Zernike moments and pseudo Zernike moments are some of the commonly used region based shape descriptors. According to the surveys [8], [3] and review [9]; the global approach for shape information extraction outperforms the structural approach. Fourier Descriptors outperform many contour based shape descriptors and Generic Fourier descriptors outperform all region based shape descriptors [9]. Centroid distance based Fourier descriptors perform better than other contour based signatures. A detailed study of several Fourier descriptors and signatures can be found in [15]. In [11] Çapar, Kurt, and Gökmen proposed a gradient based shape descriptor. Bandera in [15] proposed an adaptive approach for affine-invariant 2D shape description. As pointed out by [2] both regions based approach and contour based approach has its own pros and cons. Use of both region and contour information is the better approach [2]. Our interest is to develop an innovative descriptor as well as an algorithm that can use both regions and contour properties of shapes for object class recognition. In addition the proposed method has to be invariant to rotation, scale, orientation, affine deformations and noisy contours. Suggested integrated shape descriptor is a combination of Generic Fourier Descriptor (GFD) [1] and novel Multi Scale Gradient Based Descriptor (MSGBD). For rapid classification of object shapes, an inventive Corner Based Interpolated Descriptor (CBID) has been proposed. CBID is based on the concept proposed in [8]. In [19] a novel machine learning algorithm based on edit distance and tree matching was proposed. Similarly in [20] an approach for learning class-specific explicit shape models has been suggested. We propose to use Random Forest (RF) classifier [22] for object recognition. Similar concept in shape analysis and classification has been recommended in [23], [24], and [25].

The rest of the paper is organized as follows. The proposed method is described in Section II. Experimental Results are presented in Section III. Finally, Conclusion is presented in Section IV.

## II. METHODOLOGY

### A. Generic Fourier Descriptor

GFD is a region based shape descriptor which outperforms MPEG-7 proposed Zernike moments. As pointed out by [9], GFD has many advantages. It is easy to implement and robust. It is less sensitive to deformations and noise. GFD is invariant to rotation translation and scale. Detailed analysis of GFD is present in [9]. The authors in [1] suggest to use 9 regular and 4 angular frequencies as optimum choise for classifation. In our proposed approach $N$ number of radial lower frequency componets and $L$ number of lower angular freqeuncy components are ordered in the following format to obtain a vector.

$GFD(0,1), GFD(0,2)...GFD(0,N).GFD(1,0)...GFD(L,0)$

### B. Multi Scale Gradient Based Descriptor (MSGBD)

Gradient Based Shape Descriptor proposed by Capar and Kurt in [11] uses directional steerable filters [6]. Authors in [11] used gradient information near the contours instead of location of contour. They used G-Steerable filters to obtain the gradient information at many directions. Our proposed methodology is based on the concept of steerable filter responses at the boundary of the shape. We use canny edge detector to detect edges. Instead for *G* filters, Gaussian filters are used to obtain directional derivatives. The concept of gradient based shape descriptor [11] has been enhanced to have better description of the object. The directional filter response is computed based on the concept of steerable filters. As pointed out by Mikolajczyk and Schmid in [12] the edge response significantly varies with the scale. In order to overcome this issue we propose to use steerable filter responses computed at different scales and orientations. At a given boundary point the maximum edge response and the corresponding direction of the maximum response is taken at several scales. This information is used to create the new shape signature. Maximum edge response computed by using multiple orientations and scales is robust to affine transformations, viewpoint and noisy boundaries. The Gaussian function and its first and second major directional partial derivatives are given by

$$G(x,y) = e^{-\left(\frac{x^2+y^2}{2\sigma}\right)} \quad G_\sigma^0 = \frac{\partial}{\partial x} e^{-\left(\frac{x^2+y^2}{2\sigma}\right)} \quad G_\sigma^{90} = \frac{\partial}{\partial y} e^{-\left(\frac{x^2+y^2}{2\sigma}\right)}$$

Here the partial derivative along the x dimension is said to have directional derivative of zero degree and along y dimension its 90 degree.

Let $R_\sigma^0 = G_\sigma^0 * I$ and $R_\sigma^{90} = G_\sigma^{90} * I$

Directional filter response $R_\sigma^\theta$ and filter $G_\sigma^\theta$ at scale $\sigma$ and direction $\theta$ can be given by

$$R_\sigma^\theta = \cos(\theta) \times R_\sigma^0 + \sin(\theta) \times R_\sigma^{90} \text{ and } G_\sigma^\theta = \cos(\theta) \times G_\sigma^0 + \sin(\theta) \times G_\sigma^{90}$$

Let image be a function f(x, y); directional derivative at θ direction is defined as follows

$$g^\theta(x,y) = \left(f(x,y) * \left(\cos(\theta) \times \frac{d}{dx} G(x,y) + \sin(\theta) \times \frac{d}{dy} G(x,y)\right)\right)$$

We can write the directional filter response using the steerable filter with a given scale σ and direction θ as follows

$$g_\sigma^\theta(x,y) = f(x,y) * G_\sigma^\theta(x,y)$$

The gradient of the image *I* in multiple directions for a given scale σ is calculated as follows

$$GBD_\sigma(k,m) = I * G_\sigma^\theta(x_k, y_k)$$

Filter response magnitude $f(k)$ and directional filter response $f_d(k)$ is defined as

$$f(k) = \left(\underset{m=1}{\overset{m=M}{MAX}}(GBD_\sigma(k,m))\right) \text{ and } f_d(k) = \left(\underset{m=1}{\overset{m=M}{ArgMax}}(GBD_\sigma(k,m))\right)$$

After that for each boundary point (x, y) the polar coordinates is computed. For each boundary point (x, y) r and θ is computed. After that the proposed shape signature based on polar coordinates is expressed in the form of

$$f(r,\theta) = f + i \times f_d \quad \forall \ (x,y) \in Boundary$$

Fourier Transform is applied on this signature followed by FD normalization and takes K×L=X low frequencies as the response

$$\hat{f}(\lambda,\mu) = \sum_r \sum_\theta f(r,\theta) \times e^{-j2\pi\left(\frac{r}{R}\lambda + \frac{\theta}{T}\mu\right)}$$

The output of the MSGBD is a vector of size *X* by performing 2-D to 1-D transformation. MSGBD can encapsulate shape information in compact and robust way. MSGBD uses centroid based location information, gradient magnitude in multiple directions and scales. It's invariant to rotation, scale and affine deformations due to the fact that edge responses are quite stable over multiple scales.

*C. Integrated shape descriptor*

The proposed integrated shape descriptor combines GFD and MSGBD based on region and contour properties of shape. In order to obtain better results and to control the significance of contribution of the overall descriptor; two weighting parameters α, β are used. The combined descriptor is (N×L) + X in length one dimensional vector. So the integrated shaped descriptor ISD is written in the form

$$ISD = \alpha \times \overline{GFD} + \beta \times \overline{GBSD}$$

The distance between two shape objects is calculated based on city block distance. However for the classification task we use a Random Forest classifier.

*D. Corner based interpolated descriptor (CBID)*

Corner based interpolated descriptor is designed to find probable classes of shapes in multiclass object scenes. Being a compact descriptor, it has the capability to capture significant shape information and to rapidly classify objects into classes. According to the proposed approach, firstly the image is binarized by OTSU's method; then the Canny edge detector is applied on binarized image to obtain robust contour information. After that Harris Stephen corner detector is applied on the edge detected image to find corner points in the object boundary. Once corner points are localized; they are transformed to polar representation. Centroid of the shape object is computed using detected corner points. Suppose the set of corner points are given by

$$C = c_1(x,y), c_2(x,y), c_3(x,y) \ldots c_n(x,y)$$

Then the radial distances and the angles of each corner point is calculated as follows:

$$R_n = \sqrt{(y_n - y_c)^2 + (x_n - x_c)^2} \quad \theta_n = a\tan\left(\frac{y_n - y_c}{x_n - x_c}\right)$$

Where $(x_c, y_c)$ is the coordinates of the centroid.

Normalization for the radial distance is done by:

$$\overline{R_n} = \frac{R_n}{Max(R)}$$

and the signature is given by $f(\theta_n) = \overline{R_n}$

When we need to compare two shapes for similarity, we just need to interpolate the shape signature for every *θ* from 0 to 360 degrees using nearest neighbor interpolation. Then Fourier descriptor is applied on the interpolated shape signature. It is also possible to use other interpolation technique. For each shape class we extract the radial signatures based on the corner points. Since very few corner points can capture shape signature; this descriptor becomes very compact and it can be used with any machine learning technique. We have used Random Forest classifier but KNN or any other machine learning approach can also be used.

*E. Object Recognition and Classification*

First the corner based interpolated descriptor is applied to recognize the probable classes. It's the job of the Classifier designed for CBID to give the probability for being in each class. We select *N* number of classes with highest probability. Once probable shape classes are recognized; the actual classification is done based on the Integrated Shape Descriptor (ISD) proposed in section II (*C*). In this way the search space for ISD has been reduced. Finally in conjunction with RF classifier we have noise free classification.

### III. EXPERIMENTAL RESULTS

GFD [1] of size 9 × 4 is used for the integrated descriptor. MSGBD starts with σ = 0.1, an increment factor of 1.4 (up to 5 scale levels) and 10 directional derivatives. According our observation and analysis 36 low frequency components are sufficient to represent shape information. For experimental analysis MPEG7 CE Shape-1 Part B shape database with 70 image shape classes and 1400 images are used. Random Forest classifier used in this paper uses 10 trees; each constructed while considering 6 random features. Corner point based interpolated descriptor uses maximum of 40 corner points with nearest neighbor interpolation. To represent the shape information in CBID 10 Fourier descriptors were used. For the comparison with other algorithms we have used Centroid Based Fourier Descriptor CBFD with 36 FD coefficients. Also Elliptic Fourier Descriptor [13] EFD, GFD, CBFD, CBID and MSGBD were selected to do the comparisons. All descriptors selected used 36 coefficients except CBID (The classifier used in each shape descriptor is RF classifier). The summary of the experimental results on several shape descriptors on MPEG-7 shape dataset is as follows.

TABLE 1: EXPERIMENTAL RESULTS

| Method | GFD | EFD | CBISD | MSGBSD | CBFD |
| --- | --- | --- | --- | --- | --- |
| Accuracy | 80% | 65% | 78.33% | 85.50% | 63.33% |
| Average Precision | 0.80 | 0.64 | 0.80 | 0.86 | 0.67 |
| Average Recall | 0.80 | 0.65 | 0.78 | 0.85 | 0.63 |

### IV. CONCLUSION

Nonetheless our proposed corner based interpolated descriptor is fast and compact; it has a better accuracy than elliptic Fourier descriptors and centroid based Fourier descriptors. Corner point based interpolated descriptor performs well on largely deformed shapes and when noise is present in the contour of the shape. In this approach we used corner points however in future experiments we want to explore on visual saliency based descriptors.

Gradient information near object boundaries is very useful to extract shape information as well as region based shape properties. Since contour information varies with the scale it's very important to extract gradient information at different scales and orientations. Proposed Multi Scale Gradient Based Descriptor extracts shape information in multiple orientations and scales; it encapsulates centroid based radial information. As a result MSGBD shows very high accuracy in comparison to other shape descriptors such as GFD, CBFD. Again it is robust to noisy contours, invariant to rotation, scale and affine deformations. Finally the integrated shape descriptor is a perfect shape descriptor which captures both regions based and contour based shape properties and gives better classification with satisfactory retrieval time. It suffers when the shape object is very small, which is the only noticeable drawback of this approach. In future work we want to incorporate visual saliency information to challenge that issue.